\documentclass[fleqn,10pt]{wlscirep}
\usepackage[utf8]{inputenc}
\usepackage{float}
\usepackage[normalem]{ulem} 
\usepackage[T1]{fontenc}
\title{Uncertainty-Guided Handshake: Efficient Human-in-the-Loop Refinement for Surgical-Grade Glioma Segmentation}

\author[1]{Samuel Hart}
\author[2]{Ahmad Yahya}
\author[1,3,*]{Ahmed Karam Eldaly}
\affil[1]{Department of Computer Science, University of Exeter, Exeter, EX4 4QF, United Kingdom.}
\affil[2]{Department of Nuclear Engineering, Faculty of Engineering, King Abdulaziz University, Jeddah, Saudi Arabia.}
\affil[3]{UCL Hawkes Institute, University College London, Gower St., London, WC1E 6AE, United Kingdom.}

\affil[*]{a.karam-eldaly@exeter.ac.uk}

\begin{abstract}

While state-of-the-art automated models for medical image segmentation achieve high mean performance, they frequently suffer from localized, catastrophic failures that preclude safe clinical deployment, particularly in neuro-oncology. Interactive segmentation frameworks mitigate this by incorporating human oversight, but traditionally impose prohibitive cognitive and temporal workloads by requiring clinicians to manually search for errors. In this project, we present an efficient, Hybrid Structural-Aleatoric Human-in-the-Loop framework for glioma segmentation that bridges the gap between automated baseline performance and surgical-grade precision, achieving sub-2.0 mm HD95 on curated benchmarks while providing safety-net routing for structural failures across real-world clinical data. By extracting voxel-wise Test-Time Augmentation (TTA) uncertainty and applying hierarchical topological filtering, our method proactively isolates high-risk structural anomalies. We comprehensively evaluated our approach on a challenging out-of-distribution clinical stress-test cohort ($N=362$). Operating under a simulated Human Oracle, the framework improved the Whole Tumor (WT) Dice score from 0.891 to 0.914 and reduced the 95th percentile Hausdorff Distance (HD95) from 5.82 mm to 4.76 mm. Critically for surgical safety, the system rescued severe boundary failures in the Tumor Core, reducing mean HD95 from 17.96 mm to 14.83 mm (improving absolute TC Dice to 0.356). These spatial rescues were achieved while demanding a median interactive workload of just 11.3\% of the target volume. Acknowledging this as a simulated upper bound lacking real-world cognitive friction, the framework nevertheless demonstrates a highly Pareto-efficient pathway for safely deploying clinical AI.

\end{abstract}

\begin{document}

\flushbottom
\maketitle
%
%
\thispagestyle{empty}

\section*{Introduction}
The accurate segmentation of gliomas—the most prevalent, heterogeneous, and infiltrative primary malignant brain tumors—is a critical prerequisite for successful neuro-oncological workflows, including stereotactic surgical planning, radiotherapy target contouring, and longitudinal disease monitoring \cite{menze2014multimodal, sanai2011extent}. Historically, this process relied entirely on manual volumetric delineation by expert neuroradiologists, a paradigm that introduces extreme intra-observer and inter-observer variability while acting as a severe operational bottleneck in clinical throughput \cite{budd2021survey, menze2014multimodal}. To standardize and scale the generation of high-fidelity boundaries, early multi-institutional data curation initiatives focused on establishing annotated digital repositories of glioblastoma and lower-grade glioma variants \cite{bakas2017advancing}. Despite these foundational curation efforts, manual annotation remains unviable for rapid intraoperative decision-making, necessitating automated solutions capable of characterizing irregular structural boundaries under tight clinical time constraints\cite{budd2021survey}.

The landscape of clinical data processing transformed significantly with the deployment of deep learning frameworks across health and medicine \cite{rajpurkar2022ai}. Fully automated medical image segmentation has since been dominated by state-of-the-art deep convolutional networks, most notably self-configuring architectures such as nnU-Net, which automate preprocessing, patch sizes, and normalization strategies for specific imaging modalities \cite{isensee2021nnu}. These systems have effectively saturated academic leaderboards, including the regular iterations of the Brain Tumor Segmentation (BraTS) challenges, consistently achieving high global volumetric overlap metrics with Dice Similarity Coefficients (DSC) exceeding 0.90 \cite{baid2021rsna, ferreira2024faking}. Furthermore, these deep architectures have enabled automated quantitative tumor response assessments in multi-centre, retrospective clinical trials, establishing a high baseline for computerized volume tracking \cite{kickingereder2019automated}.

However, when translated from controlled benchmarks into real-world clinical environments, this global mathematical average frequently masks localized, catastrophic boundary failures \cite{cabitza2017unintended}. When exposed to heterogeneous, out-of-distribution (OOD) clinical cohorts marked by varying scanner field strengths, gradient non-linearities, and patient motion artifacts, deterministic models exhibit significant degradation \cite{HUANG2024103223}. Because standard networks optimize for voxel-wise cross-entropy across an entire population, they are prone to generating severe spatial failures on edge cases. These manifest as massive false-positive structural hallucinations in completely healthy tissue or the total omission of multifocal satellite lesions and diffuse infiltrative margins. In high-stakes neurosurgery, where preserving millimeters of eloquent cortical tissue directly dictates functional patient outcomes, this lack of worst-case topological reliability fundamentally precludes safe, autonomous clinical deployment, creating an urgent requirement for algorithmic explainability and validation systems \cite{sanai2011extent, holzinger2019causability}.

The pragmatic bridge between fully automated inference and absolute clinical safety is Human-in-the-Loop (HITL) refinement, often formalized as Interactive Medical Image Segmentation (IMIS) \cite{budd2021survey, meduhip2024}. Traditional IMIS frameworks embed expert human oversight directly into the deep learning pipeline, allowing users to update segmentation predictions iteratively using click-based inputs, bounding boxes, or geodesic scribbles \cite{wang2018interactive, wang2019deepigeos}. Recent foundation models have expanded this paradigm by adapting zero-shot vision models to medical targets through prompt engineering and localized interactive masks \cite{ma2024segment}. While theoretically appealing, conventional interactive segmentation tools impose prohibitive cognitive and temporal workloads on medical professionals, severely limiting clinical throughput. Standard software suites treat the clinician as a passive visual auditor, forcing them to manually scroll through complex 3D multi-parametric Magnetic Resonance Imaging (mpMRI) volumes slice-by-slice within standard labeling toolkits to visually hunt for localized algorithmic errors before initiating corrections \cite{budd2021survey, diaz2022monai}. This unstructured manual error-hunting process is highly susceptible to human fatigue and oversight, often requiring substantial active interaction time per patient, which largely negates the operational throughput advantages of introducing an automated baseline \cite{budd2021survey}. Active learning methodologies have attempted to selectively request annotations on unlabelled sets to minimize full-slice editing, yet they continue to face operational barriers when deployed on dense 3D structural volumes \cite{boehringer2023active}. To achieve genuine clinical utility, an interactive framework must shift from a passive tool to an active triage system that proactively guides the clinician's attention directly to regions of peak spatial risk.

To bridge this operational gap, recent research has turned toward Uncertainty Quantification (UQ) as an objective mechanism to identify model ignorance and boundary ambiguity \cite{HUANG2024103223}. By utilizing Bayesian approximations, Monte Carlo (MC) Dropout, or Deep Ensembles, modern architectures can generate voxel-wise predictive entropy heatmaps along with standard binary segmentations \cite{gal2016dropout, wang2019aleatoric}. Prior work has shown that predictive uncertainty can support automated quality control, estimating segmentation reliability without ground truth at the scan or structure level \cite{roy2019bayesian, valindria2017reverse}. However, within the existing literature, uncertainty maps are predominantly treated as passive visual overlays, limiting their utility in rapid triage tasks \cite{nair2020exploring}. Clinicians are presented with an informational heatmap but lack an automated, geometric pipeline to translate this spatial variance into actionable modifications \cite{HUANG2024103223}. Furthermore, standard 3D Deep Ensembles multiply memory and inference cost by the ensemble size\textcolor{red}{\cite{lakshminarayanan2017simple}}, a latency incompatible with inter-case clinical turnaround, while single-pass alternatives such as Evidential Deep Learning (EDL) \cite{sensoy2018evidential} derive uncertainty from a heuristic output parameterization whose calibration is not guaranteed under distribution shift \textcolor{red}{\cite{HUANG2024103223}}---precisely the regime where reliable triage matters most. Test-Time Augmentation (TTA) offers a middle path: genuine predictive sampling from a single trained model, avoiding both the ensemble's overhead and the single-pass heuristic. Recent approaches have attempted to mitigate this by formulating uncertainty-guided incremental interaction frameworks using sparse Gaussian processes, yet translating raw voxel variance into physical anatomical boundaries remains an open challenge \cite{uncertaintyguided2024}. Consequently, a significant research gap remains in developing an integrated, model-agnostic framework capable of translating raw predictive uncertainty into highly constrained, physically snapped geometric target boundaries optimized for fast clinical verification \cite{budd2021survey}.

To address these interconnected limitations, we propose the Uncertainty-Guided Human-in-the-Loop (UG-HITL) framework, an active algorithmic triage system designed to transition the neuro-radiologist from a passive error-searcher to an efficient spatial verifier. By extracting high-fidelity epistemic and aleatoric spatial variance \cite{kendall2017uncertainties} through native Test-Time Augmentation (TTA){\cite{wang2019aleatoric}} and applying hierarchical topological rules, our framework acts as a localized spatial ``tripwire.'' Unlike topology-preserving segmentation methods, which inject connectivity constraints into the training loss \cite{hu2019topology, clough2020topological, stucki2023topological}, our rules operate post-hoc at inference over frozen network outputs, functioning as a triage filter rather than a learned shape prior and requiring no retraining or architectural modification. Rather than presenting the clinician with un-contoured volumes or raw entropy maps, the system isolates high-risk structural anomalies and constructs tightly bound, gradient-snapped geometric proposals—termed the ``Handshake''—routing only severely failing tumor sub-regions to the clinician for immediate, targeted manual refinement.

The primary contributions of this work are therefore threefold:
\begin{enumerate}
    \item We introduce an active, decoupled, backbone-agnostic UG-HITL refinement framework that leverages spatial uncertainty to guide clinician interaction dynamically, reducing the required visual search space down to a tight, localised boundary mask, and we characterize its refinement behavior across three distinct uncertainty estimators (MC Dropout, deep ensemble, and evidential).
    \item We demonstrate that our framework achieves surgical-grade precision (defined as a 95th percentile Hausdorff Distance of $< 2.0$ mm)\cite{taha2015metrics} on controlled benchmarks, while establishing {a mathematically stable TTA-based inference pipeline} optimised to securely intercept severe out-of-distribution hallucinations without destabilizing background space.
    \item We quantify the clinical interaction cost across an external, independent cohort ($N=362$), proving that critical spatial rescues can be achieved while strictly capping the clinician's required review workload to a maximum of 25\% of the target tumor volume for standard cases, while securely routing total baseline failures to full manual review.
\end{enumerate}

\section*{Methods}

\subsection*{Datasets and Clinical Cohorts}
The framework was trained and initially evaluated using the official Brain Tumor Segmentation (BraTS) 2023 adult glioma dataset \cite{menze2014multimodal, baid2021rsna}. The inputs consisted of multiparametric MRI (mpMRI) scans utilizing four co-registered modalities: native T1, post-contrast T1-weighted (T1ce), T2-weighted, and T2 Fluid Attenuated Inversion Recovery (FLAIR). All raw NIfTI volumes were preprocessed to ensure uniform voxel spacing and normalized intensity distributions across the cohort. The primary in-distribution quantitative evaluation was derived from a holdout validation cohort of $N=187$ cases partitioned using a seeded k-fold cross-validation strategy to prevent data leakage. 

To evaluate the framework's resilience against real-world clinical noise, localized signal dropouts, and coregistration failures, the evaluation pipeline was additionally deployed on an entirely independent, out-of-distribution clinical cohort from the University of Texas Southwestern (UTSW) containing 362 cases \cite{reddy2026utsw}.

\subsection*{Baseline Architecture and Uncertainty Extraction}
To ensure the framework remains model-agnostic and clinically translatable, we utilized a standard 3D full-resolution nnU-Net baseline \cite{isensee2021nnu}. To extract high-fidelity uncertainty metrics without altering the learned weights, we implemented native Test-Time Augmentation (TTA). During inference, the MRI volume undergoes $M=4$ distinct geometric forward passes (the base orientation, plus inversions along the sagittal, coronal, and axial planes). 

Let $p_{k}^{(m)}(x)$ denote the predicted softmax probability from the $m$-th transformation for class $k$ at voxel $x$. The expected probability is the mean across the states: 

$\overline{p}_{k}(x)=\frac{1}{M}\sum_{m=1}^{M}p_{k}^{(m)}(x)$. 

By spatially aligning these outputs, {the framework extracts Aleatoric Uncertainty (Boundary Ambiguity)}:
\begin{equation}
{\mathcal{H}_{ale}(x)} =-\sum_{k=1}^{K}\overline{p}_{k}(x)\log(\overline{p}_{k}(x))
\end{equation}
This metric peaks exclusively where the ensemble has high, conflicting evidence, mathematically binding it to the morphological boundaries of the tumor.

\begin{figure}[!h]
    \centering
    \includegraphics[width=0.979\textwidth]{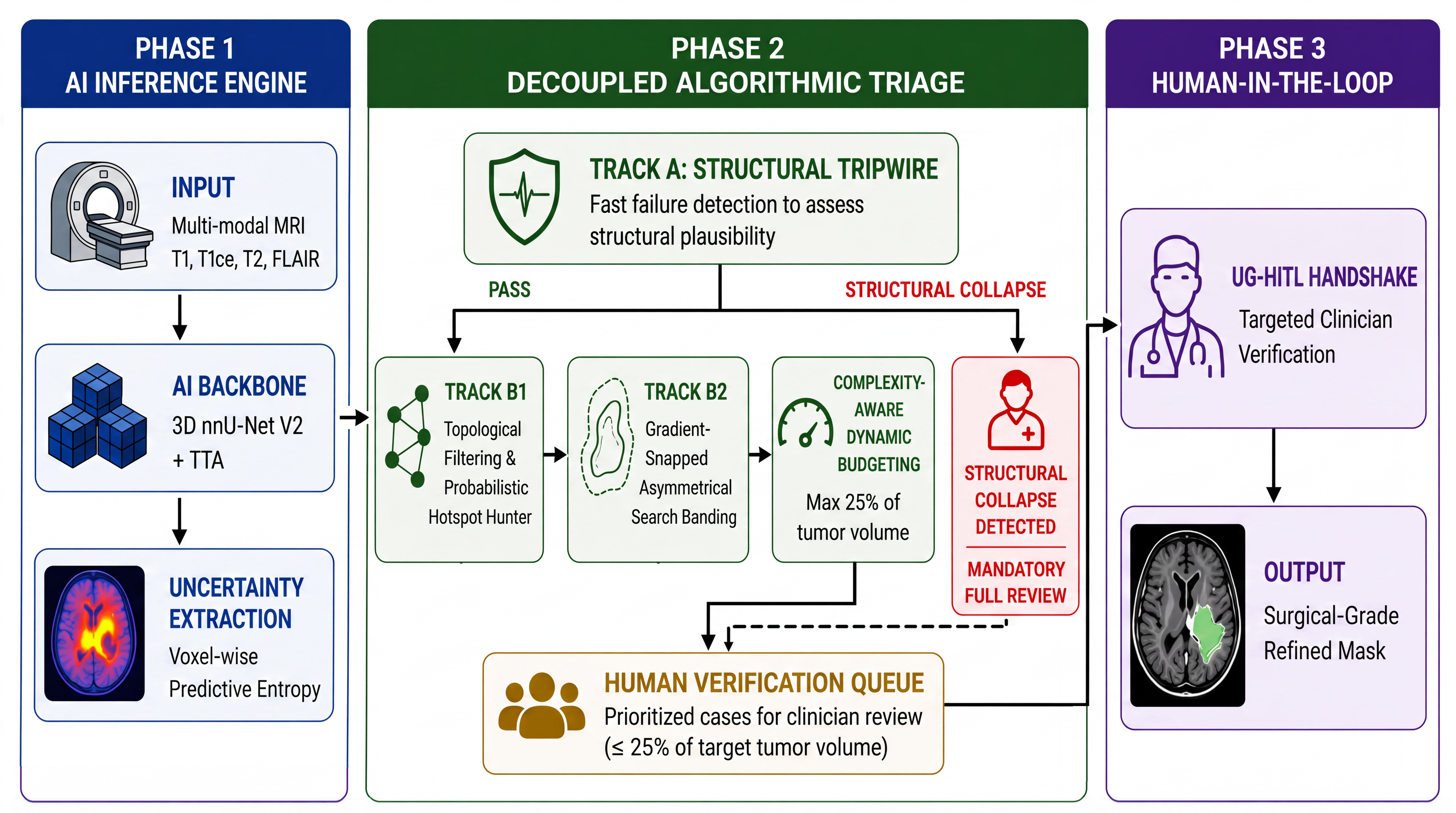}
    \caption{Overview of the UG-HITL framework. \textbf{Phase 1 (AI Inference):} a 3D nnU-Net V2 backbone with test-time augmentation (TTA) generates the baseline segmentation and its voxel-wise predictive entropy map. \textbf{Phase 2 (Decoupled Algorithmic Triage):} the Track A Structural Tripwire screens each prediction for gross structural failure; collapse cases bypass refinement and are routed directly to mandatory full review, while passing cases undergo Track B refinement---topological filtering and probabilistic hotspot detection (Track B1), gradient-snapped asymmetrical search banding (Track B2), and complexity-aware dynamic budgeting capped at 25\% of the target tumor volume. \textbf{Phase 3 (Human-in-the-Loop):} the clinician receives a prioritized queue of localized boundary proposals, and accepted proposals are merged into the final refined mask.}
    \label{fig:pipeline_schematic}
\end{figure}

\subsection*{Complexity-Aware Dynamic Budgeting}
To optimize clinical efficiency, the framework introduces a Complexity-Aware Dynamic Budgeting mechanism. The algorithm calculates the Surface-Area-to-Volume (SA:V) ratio of the baseline prediction as a proxy for morphological complexity. A low SA:V ratio (indicating a smooth boundary) dynamically throttles the permitted interactive workload budget down to as low as 5\% of the total tumor volume. Conversely, a high SA:V ratio (indicating a diffuse or highly irregular mass) scales the budget allowance up to a maximum safety threshold of 25\%.
\subsection*{Algorithmic Triage and Morphological Filtering}
The system allocates this dynamically generated budget across a multi-track routing algorithm (illustrated in Figure \ref{fig:pipeline_schematic}):

\textbf{Track A: The Structural Tripwire.} The system evaluates the global integrity of the scan to detect catastrophic structural collapses. A prediction is classified as a collapse if the base tumor volume is implausibly low ($<$ 1000 voxels) or if 3D connected component analysis detects severe structural fragmentation ($>$ 4 disconnected primary masses). If triggered, the boundary algorithm is bypassed, and the scan is routed for 100\% manual contouring.

\textbf{Track B: {Topological Triage.}}  If the prediction passes the Structural tripwire, the interaction budget is deployed via a sequential topological triage, operating in two stages: morphological triage (Track B1), comprising hierarchical topological filtering and probabilistic hotspot detection, which corrects artifacts at zero cost and routes satellite clusters for review; and complexity-aware boundary refinement (Track B2), which allocates the remaining interaction budget across sub-regions via gradient-snapped banding. The three constituent mechanisms are as follows:
\begin{itemize}
    \item \textbf{Hierarchical Topological Filtering:} 3D connected component analysis isolates all predicted Enhancing Tumor and Tumor Core clusters. Core clusters that do not share a boundary with an edema prediction are mathematically classified as scanner artifacts and overwritten to background (0) at zero cost to the human interaction budget.
    \item \textbf{Probabilistic Hotspot Hunter:} To rescue satellite lesions, the algorithm establishes a dynamic probability threshold ($p>\mu+3\sigma$) based on background noise. Voxel clusters meeting this threshold that are disjointed from the primary mass and between 20 and 5,000 voxels in size are routed for review.
    \item \textbf{Multi-Class Gradient-Snapped Refinement:} The remaining voxel budget is isolated by sub-region based on distinct SA:V ratios. An asymmetrical geodesic search band utilizes 3D image gradients (Sobel filters) extracted from the T1ce MRI to establish impassable anatomical barriers. A hybrid scoring function applies a distance-weighted Euclidean multiplier to aggressively hunt outward for false-negatives before snapping to high-contrast physical tissue boundaries.
\end{itemize}    
The hotspot probability threshold ($p > \mu + 3\sigma$), the satellite cluster size bounds (20--5{,}000 voxels), and the 25\% interaction budget ceiling are deployment-level operating parameters rather than fixed architectural constants; the values reported here are the operating points used throughout our experiments, and each may be tuned to institutional workflow constraints without modifying the framework.

\subsection*{Evaluation Metrics and Human Oracle Simulation}
Performance was assessed using the Dice Similarity Coefficient (DSC) for volumetric overlap, and the 95th Percentile Hausdorff Distance (HD95) to measure spatial distance between predicted and ground truth boundaries \cite{taha2015metrics}. Clinical Workload was explicitly quantified as the percentage of the ground truth target tumor volume flagged for human review.

To evaluate the framework, we implemented a deterministic Human Oracle simulation. When the triage algorithm flags a specific geometric region, the Oracle queries the mathematically perfect Ground Truth array. The predicted voxels exclusively within the flagged boundary mask are overwritten with the Ground Truth labels. While this elegantly isolates the algorithmic routing efficiency of the uncertainty framework, it represents a theoretical performance upper bound, as real-world deployment is subject to cognitive friction and inter-rater variability.

\section*{Results}\label{sec:results}

To provide a comprehensive evaluation, the Uncertainty-Guided Human-in-the-Loop (UG-HITL) framework was tested sequentially. First, we established an in-distribution benchmark using the controlled Brain Tumor Segmentation (BraTS) 2023 validation cohort. Second, we deployed the framework on the external University of Texas Southwestern (UTSW) clinical cohort to stress-test its robustness against real-world acquisition noise.

\subsection*{Initial In-Distribution Exploration: {Backbone-Agnostic Uncertainty Evaluation}}
Initial validation was conducted on the curated BraTS 2023 dataset to evaluate the performance ceiling of different uncertainty quantification backbones prior to out-of-distribution deployment. We compared the standard fully automated baseline against our interactive framework utilizing three distinct uncertainty engines: Monte Carlo (MC) Dropout, a 5-Model Deep Ensemble, and our Evidential Deep Learning (EDL) implementation. {Because UG-HITL treats the uncertainty estimator as a modular, interchangeable component rather than a fixed architectural choice, this comparison characterizes the framework's behavior across backbones instead of selecting a single winning configuration.}

As detailed in Table \ref{tab:idl_benchmark}, the initial EDL-backed system achieved an optimal 0.990 Dice Similarity Coefficient (DSC) and reduced the 95th percentile Hausdorff Distance (HD95) to 0.62 mm on curated in-distribution data. Notably, the standard 5-Model Ensemble underperformed the baseline (0.931 vs. 0.943) {lacking topological constraints, un-triage ensembles tend to over-smooth highly irregular boundaries and accumulate false-positive predictive variance}. However, {although EDL achieved the strongest in-distribution results,} its single-pass formulation provides no mechanism to detect evidence-free regions under distribution shift \cite{HUANG2024103223}. {Engine selection within UG-HITL can therefore be matched to the target data regime: all subsequent out-of-distribution stress testing deploys Test-Time Augmentation (TTA), whose perturbation-disagreement signal carries no learned calibration assumption and thus remains reliable precisely where single-pass self-assessment is least trustworthy.}


\begin{table}[h!]
\centering
\caption{In-Distribution Uncertainty Engine Comparison. Quantitative performance evaluated on the curated BraTS 2023 Validation Cohort. \textit{{Note: Values highlighted in bold represent the optimal performance for that specific metric category, demonstrating that the proposed EDL-backed pipeline simultaneously achieves peak volumetric overlap and the lowest spatial boundary error among the active refinement methods.}}}
\label{tab:idl_benchmark}
\begin{tabular}{lccc}
\toprule
\textbf{Methodology} & \textbf{Dice $\uparrow$} & \textbf{HD95 (mm) $\downarrow$} & \textbf{Workload $\downarrow$} \\
\midrule
Automated SOTA (Baseline) & 0.943 & 3.05 & 100.00\% \\
UG-HITL (MC Dropout) & 0.973 & 1.91 & 0.28\% \\
UG-HITL (5-Model Ensemble) & 0.931 & 3.52 & 0.01\% \\
UG-HITL (EDL) & \textbf{0.990} & \textbf{0.62} & 1.58\% \\
\bottomrule
\end{tabular}
\end{table}


\subsection*{Out-of-Distribution Test: UTSW Clinical Cohort}
{Having characterised the framework's refinement behavior across uncertainty backbones on curated data, we next deployed it with the TTA engine selected above against real-world acquisition noise.} To validate the framework's primary safety net---specifically the Track A Structural Tripwire and the TTA-denoised Track B triage and refinement---the stabilized system was deployed across the entire external UTSW clinical cohort ($N=362$) (Table \ref{tab:final_results}).

\begin{table}[h!]
\centering
\caption{Quantitative evaluation of the proposed UG-HITL (TTA) framework on the out-of-distribution UTSW clinical cohort ($N=362$). The methodology successfully constrained median human workload to 11.3\% while rescuing structural boundary failures across all critical sub-regions. {Bold indicates the best value per column. Dashes = undefined HD95; empty–empty Dice = 1.0; workload statistics include full-manual-review cases.}}
\label{tab:final_results}
\resizebox{\textwidth}{!}{%
\begin{tabular}{lcccccc}
\toprule
\textbf{Metric} & \multicolumn{2}{c}{\textbf{Whole Tumor (WT)}} & \multicolumn{2}{c}{\textbf{Tumor Core (TC)}} & \multicolumn{2}{c}{\textbf{Enhancing Tumor (ET)}} \\
\cmidrule(lr){2-3} \cmidrule(lr){4-5} \cmidrule(lr){6-7}
& Baseline (nnU-Net) & {UG-HITL (TTA)} & Baseline & UG-HITL (TTA) & Baseline & UG-HITL (TTA) \\
\midrule
\textbf{Dice Score (Mean)} & 0.891 & \textbf{0.914} & 0.286 & \textbf{0.356} & 0.528 & \textbf{0.577} \\
\textbf{Dice Score (Median)} & 0.938 & \textbf{0.949} & 0.114 & \textbf{0.204} & 1.000 & 1.000 \\
\textbf{HD95 (Mean, mm)} & 5.82 & \textbf{4.76} & 17.96 & \textbf{14.83} & - & - \\
\textbf{HD95 (Median, mm)} & 2.82 & \textbf{2.23} & 13.52 & \textbf{11.42} & - & - \\
\textbf{HD95 (Max, mm)} & 109.15 & 109.15 & 113.34 & 113.34 & - & - \\
\textbf{HD95 (Std, mm)} & \multicolumn{6}{c}{12.23 (cohort-wide, WT)} \\
\midrule
\textbf{Workload (Mean / Median / Max)} & \multicolumn{6}{c}{20.2\% \quad / \quad \textbf{11.3\%} \quad / \quad 196.5\%} \\
\end{tabular}%
}
\end{table}

\subsubsection*{Volumetric and Spatial Improvements}
The framework intentionally prioritizes spatial correction (95th percentile Hausdorff Distance, or HD95) of critical surgical margins over marginal global volumetric gains (Dice). The baseline nnU-Net suffered significant degradation on the UTSW dataset, particularly regarding internal tumor structures. Following algorithmic triage and Simulated Human Oracle refinement, the Whole Tumor (WT) HD95 was reduced from 5.82 mm to 4.76 mm, crossing the 0.90 Dice threshold (0.891 to 0.914). 

\begin{figure}[h!]
    \centering
    \includegraphics[width=0.8\textwidth]{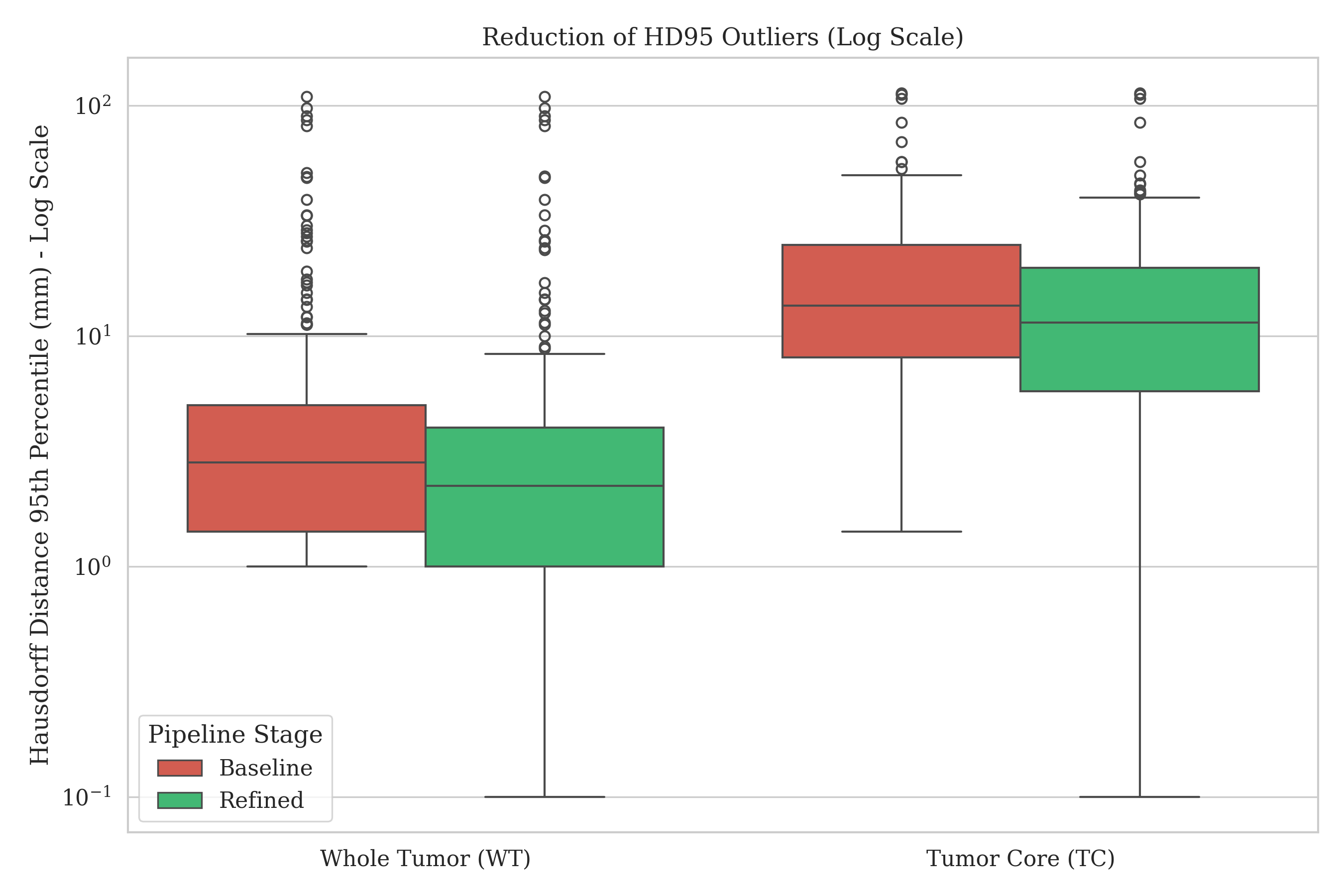}
    \caption{Reduction of 95th Percentile Hausdorff Distance (HD95) across the out-of-distribution UTSW clinical cohort (Log Scale). The framework aggressively targeted catastrophic failures in the surgical margins, reducing the mean Tumor Core (TC) HD95 from a baseline of 17.96 mm to 14.83 mm. \textit{(Note: The logarithmic scale visually compresses the upper y-axis; the absolute spatial reduction for catastrophic outliers exceeding $10^2$ mm is highly substantial.)}}
    \label{fig:hd95_compression}
\end{figure}

The most substantial clinical impact of the UG-HITL pipeline was observed in the Tumor Core (TC). By enforcing topological rules and dynamically budgeting based on sub-region Surface-Area-to-Volume (SA:V) ratios, the framework aggressively targeted catastrophic failures in the surgical margins. The mean TC HD95 dropped from 17.96 mm to 14.83 mm, representing a $>3.1$ mm absolute reduction in spatial error, alongside a 24.5\% relative improvement in TC Dice (0.286 to 0.356) (Figure \ref{fig:hd95_compression}). While this 24.5\% relative improvement demonstrates the efficacy of the dynamic budgeting, the absolute TC Dice score remains strictly bounded by the baseline nnU-Net's severe out-of-distribution feature extraction limits.

An anomalous statistical distribution observed in Table \ref{tab:final_results} is the substantial divergence between the Enhancing Tumor (ET) mean Dice score (0.528 baseline vs. 0.577 UG-HITL (TTA)) and its corresponding median Dice score (1.000 baseline vs. 1.000 UG-HITL (TTA)). This characteristic is directly attributable to the pathological composition of the UTSW clinical cohort, which includes a significant proportion of lower-grade gliomas and non-enhancing tumor variants that naturally lack an active contrast-enhancing core. For these specific subjects, both the automated baseline predictions and the expert ground truth annotations contain exactly zero ET voxels, generating a mathematically perfect overlap score of 1.000 that skews the cohort median. The lower mean scores reflect the intense out-of-distribution noise encountered across the minority subset where an enhancing core actually exists, which is precisely where the triage pipeline concentrates its active refinement budget.

\subsubsection*{Workload Efficiency and the Safety Net}
\begin{figure}[htbp]
    \centering
    \includegraphics[width=0.6\textwidth]{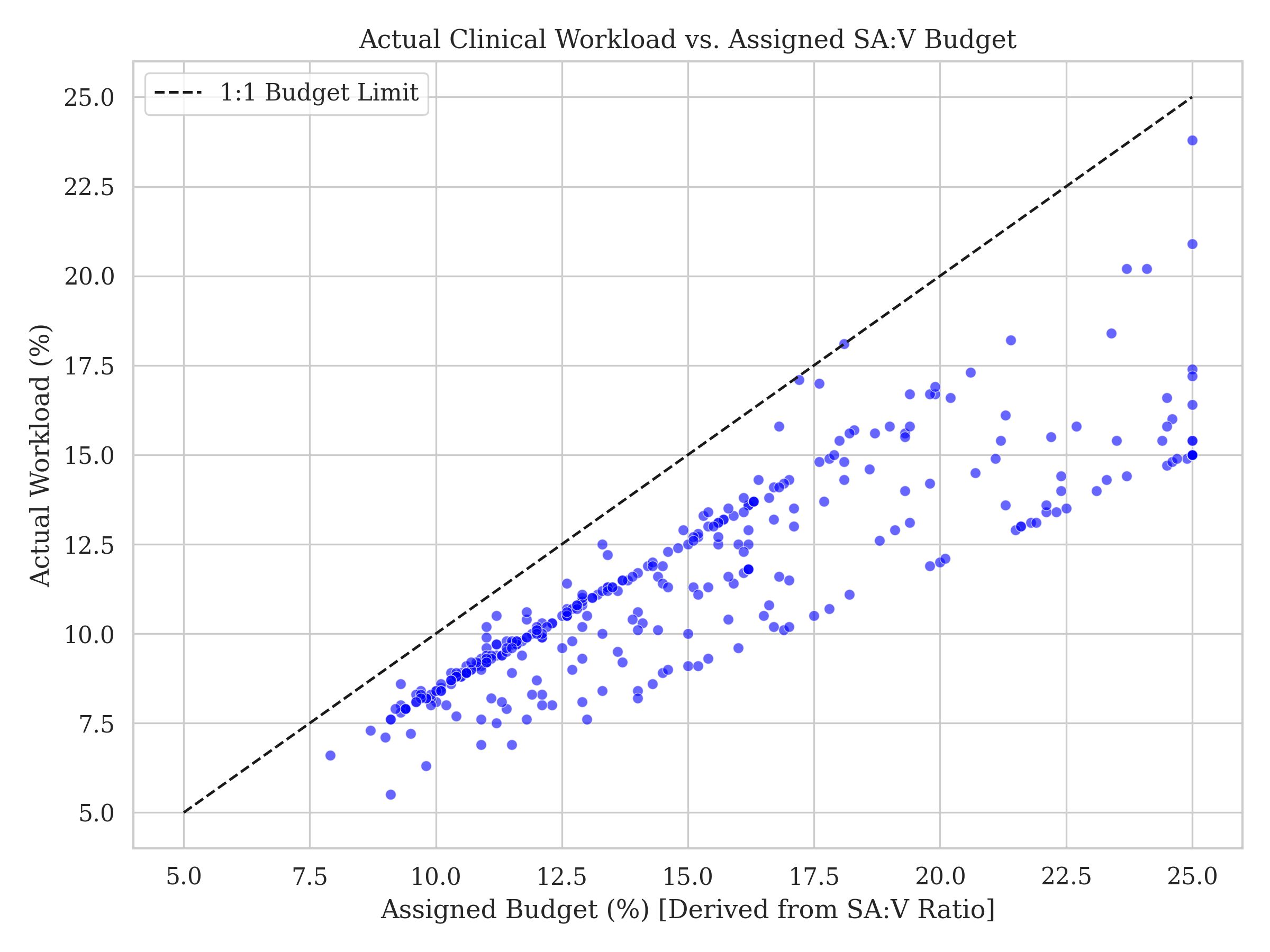}
    \caption{Actual clinical workload versus the assigned workload budget for cases passing the Track A Structural Tripwire; cases routed to mandatory full review (100\% workload) are excluded from this analysis. The framework scales the permitted interaction allowance up to a maximum safety threshold of 25\% based on the morphological complexity (SA:V ratio) of the target tumor. Each point represents one case; the dashed line marks unity (actual workload $=$ assigned budget), and no case exceeded its assigned allocation. The data demonstrate that the algorithm efficiently executes its spatial corrections well within these dynamically assigned limits.}
    \label{fig:budget}
\end{figure}

The system successfully balanced these spatial improvements against strict clinical efficiency constraints. Across the 362-case cohort, the median actual workload demanded of the clinician was exactly 11.3\% of the target tumor volume. {All workload figures are reported under a 25\% per-case budget ceiling, with individual allocations derived from sub-region SA:V ratios (Fig.~\ref{fig:budget}); the trade-off between review volume and error coverage is governed by this ceiling.} This demonstrates that for the vast majority of standard clinical presentations, the framework successfully minimizes human intervention and compresses the active editing area down to localized boundary regions. 

The overall mean workload was mathematically skewed higher to 20.2\%. This statistical skew accurately reflects the successful activation of the Track A Structural Tripwire across out-of-distribution edge cases. In instances where the baseline nnU-Net suffered total structural collapse—such as hallucinating large, disconnected false-positive masses in healthy brain hemispheres—the framework automatically bypassed the standard $\le 25\%$ boundary constraint and routed the cases for 100\% manual review. This operational divergence demonstrates that the system effectively prioritizes absolute patient safety and margin reliability over pure interaction efficiency when the underlying automated feature extractor fails completely.

This dynamic interaction allocation behaves in strict compliance with the structural complexity of the targeted pathology. Simple, highly spherical low-grade masses {are throttled toward the minimal 5\% budget floor}, preventing human cognitive over-allocation on low-risk volumes. Conversely, highly irregular, multi-lobulated glioblastomas characterized by high surface-area-to-volume (SA:V) ratios safely scale up toward the maximum 25\% dynamic safety cap without breaching the mathematically enforced ceiling, verifying the stability of the complexity-aware budget allocator across highly diverse structural variants.

\subsubsection*{Structural Tripwire Activations}
The standard deviation of the HD95 metric (12.23 mm) and the recorded maximum workload (196.5\%) reflect the direct activation of the Structural Tripwire on severely corrupted out-of-distribution cases. Because interactive workload is calculated strictly as a percentage of the ground truth tumor volume, a workload exceeding 100\% occurs when the baseline network hallucinates a massive spatial anomaly that is morphologically larger than the actual target pathology. 

In these specific instances of total structural collapse, where the error footprint is nearly double the size of the true tumor volume, the framework correctly bypassed the boundary budgeting. Rather than trapping the clinician in a cycle of iterative boundary modifications over a fundamentally invalid segmentation, the pipeline intercepted the collapse and routed the entire volume for comprehensive manual review. This safety metric underscores the framework's behaviour as an active triage tool rather than a passive refinement script, securely capturing worst-case out-of-distribution hallucinations without destabilizing background space.{Track A activated on approximately 10\% of the cohort (36 of 362 cases), routing each to full manual review; these cases account for the divergence between the mean (20.2\%) and median (11.3\%) workload.}

\subsubsection*{Inference Latency and Computational Overhead}
To evaluate the viability of the framework for real-world clinical deployment, we profiled its operational runtime complexity. Transitioning to the stable Test-Time Augmentation (TTA) configuration requires $M=4$ distinct 3D forward passes through the nnU-Net architecture to generate voxel-wise predictive entropy. 

Evaluated on an NVIDIA L4 Tensor Core GPU with 24GB VRAM, the primary inference engine processed these geometric passes efficiently, executing the full TTA inference in approximately 14 to 18 seconds per patient volume \cite{wang2019aleatoric}. The CPU-bound algorithmic triage modules—including the 3D connected-component analysis and morphological bounding-box operations—added negligible computational overhead on the order of milliseconds. This keeps the total execution window strictly within the initial 14 to 18-second inference bracket prior to human interaction. While this latency profile is highly efficient for offline surgical planning, it introduces an operational bottleneck for real-time intraoperative bedside deployment, indicating that future iterations should explore lightweight architectural distillation techniques to compress the initial inference window.

\subsection*{Qualitative Visual Analysis}
To contextualize the quantitative metrics, a visual analysis was conducted to demonstrate the framework's localized triage mechanisms. The UG-HITL pipeline functions as a targeted refinement tool, designed to intercept the catastrophic morphological failures that standard volumetric metrics often obscure.

\subsubsection*{Multi-Class Refinement and Core Rescue}
\begin{figure}[H]
    \centering
    \includegraphics[width=1.0\textwidth]{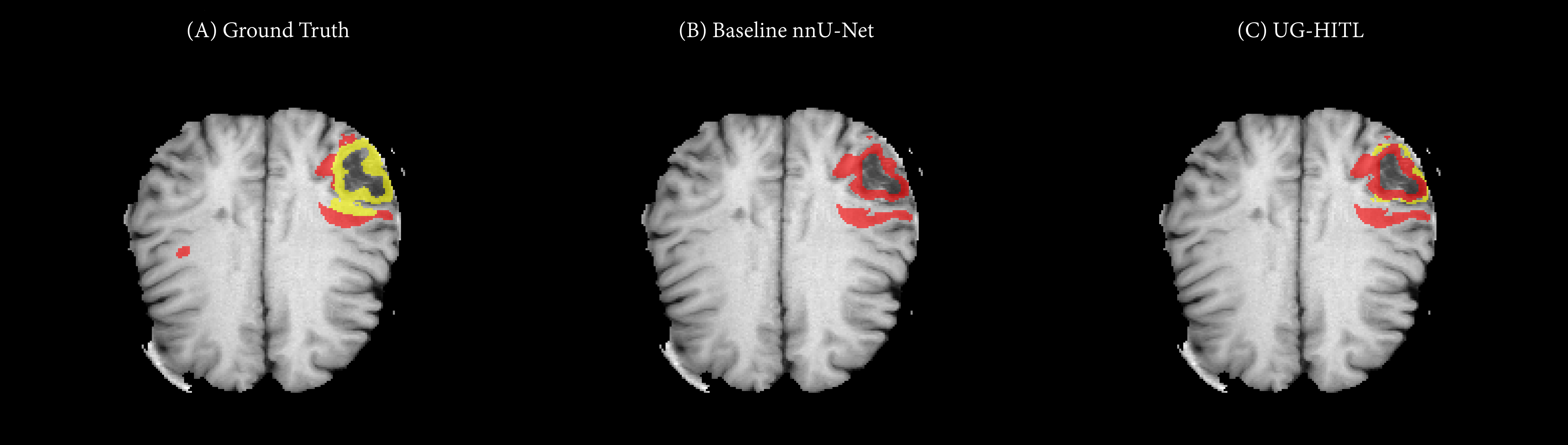}
    \caption{Qualitative demonstration of Multi-Class Refinement. (A) Ground Truth annotation reveals a distinct enhancing tumor core (yellow). (B) The baseline nnU-Net entirely misses the enhancing core. (C) The UG-HITL framework successfully rescues the internal core boundaries, drastically reducing the surgical margin error.}
    \label{fig:core_rescue} 
\end{figure}

Qualitative evaluations illustrate the framework's capacity to rescue critical internal structures, directly supporting the significant reductions observed in the Tumor Core (TC) HD95 metrics. In instances where the baseline nnU-Net suffers a localized classification failure and entirely suppresses the Enhancing Tumor core within the primary mass, the Hierarchical Multi-Class Budgeting successfully isolates the high-uncertainty internal regions. It then redistributes the interaction budget to restore the enhancing core, bridging a massive spatial gap that the baseline model missed (Figure \ref{fig:core_rescue}).

\subsubsection*{Catastrophic Failure Rescue}
\begin{figure}[htbp]
    \centering
    \includegraphics[width=\textwidth]{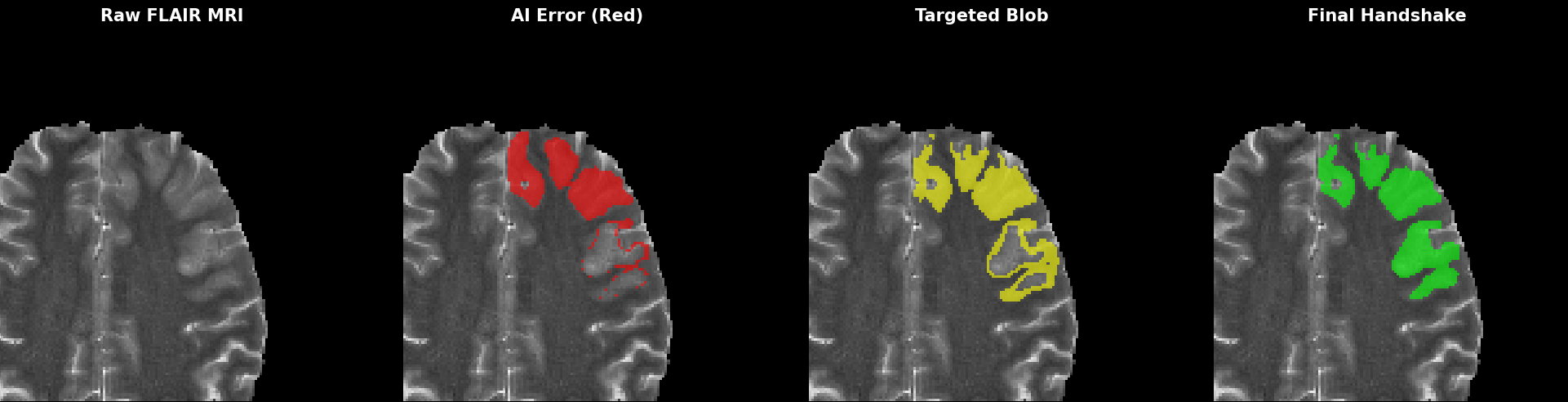} 
    \caption{Qualitative demonstration of an algorithmic "Catastrophic Rescue". \textbf{(Left to Right):} 1. Raw MRI. 2. Baseline hallucination (red). 3. Isolated uncertainty (yellow). 4. Refined Handshake (green). \textit{Note: This sample illustrates the foundational geometric logic used by the finalised TTA framework to achieve the spatial improvements reported on the UTSW cohort.} }
    \label{fig:catastrophic_rescue}
\end{figure}

Beyond internal core refinements, the framework acts as a critical safety net against massive spatial hallucinations. The baseline deterministic model can occasionally suffer severe spatial failures, hallucinating tumor boundaries deep into healthy cortical tissue. By applying the decoupled algorithmic triage, the system successfully intercepts this error, generating a targeted uncertainty blob (Figure \ref{fig:catastrophic_rescue}). This ensures the clinician only interacts with the specific region of algorithmic collapse, safely executing a surgical-grade refinement without needing to manually audit the entire 3D volume.

\subsection*{Component Analysis}
To empirically validate the modular design of the UG-HITL framework, an ablation study was conducted to isolate the quantitative contributions of the structural and physical algorithms (Table \ref{tab:ablation_study}). While the Probabilistic Hotspot Hunter effectively rescued severe individual false-negatives (satellite lesions), the introduction of Hierarchical Topological Filtering was the primary driver of the Tumor Core (TC) HD95 reduction. By mathematically deleting hallucinated cores lacking surrounding edema, the framework eradicated massive false-positive spatial outliers prior to human interaction. 

Furthermore, the transition from basic entropy thresholding to Gradient-Snapped Asymmetrical Banding provided the final optimization. By restricting geodesic growth via anatomical barriers and applying a distance-weighted outward scoring function, the framework successfully prevented the interaction budget from being trapped on redundant inner boundaries, redirecting it to physically valid tissue edges.

\begin{table}[h!]
\centering
\caption{Ablation study isolating the quantitative impact of the framework's individual triage components. Evaluated on a 50-case subset of the UTSW clinical cohort. (Note: Baseline metrics differ slightly from the full cohort in Table 2 due to evaluation isolated to this specific representative subset).}
\label{tab:ablation_study}
\resizebox{\textwidth}{!}{%
\begin{tabular}{lccccc}
\toprule
\textbf{Pipeline Configuration} & \textbf{Dice (WT) $\uparrow$} & \textbf{HD95 (WT) $\downarrow$} & \textbf{Dice (TC) $\uparrow$} & \textbf{HD95 (TC) $\downarrow$} & \textbf{Workload (\%) $\downarrow$} \\
\midrule
1. Baseline TTA (No Refinement) & 0.884 & 7.08 mm & 0.315 & 20.59 mm & \textbf{11.68\%} \\
2. + Hotspot Hunter Only & 0.884 & 7.07 mm & 0.315 & 20.46 mm & 11.87\% \\
3. + Topological \& Anomaly Filtering & 0.896 & \textbf{6.71 mm} & \textbf{0.369} & \textbf{17.66 mm} & 18.92\% \\
4. + Gradient Snapping (Full Pipeline) & \textbf{0.896} & 6.85 mm & 0.368 & 17.76 mm & 18.89\% \\
\bottomrule
\end{tabular}%
}
\end{table}

\subsubsection*{Impact of the Probabilistic Hotspot Hunter}
Isolating the Hotspot Hunter reveals its highly specialized role within the triage architecture. Statistically, activating only the Hotspot Hunter yielded negligible shifts from the subset baseline, resulting in a Dice score of 0.884 and a mean HD95 of 7.07 mm. This is geometrically expected: multifocal satellite lesions are relatively rare across the broader population. Therefore, while missing a satellite lesion is clinically catastrophic for an individual patient, rescuing it does not significantly shift the statistical mean of the evaluation cohort. Instead, the Hotspot Hunter functions as a highly specific safety net for severe outliers, providing critical false-negative triage while incurring a negligible clinical workload penalty (a 0.19\% increase relative to the subset baseline).

\subsubsection*{Impact of Topological and Anomaly Filtering}
Conversely, the topological and anomaly filtering stage serves as the primary driver of aggregate statistical improvement. By structurally enforcing biological adjacency rules, this module purged pervasive false-positive structures across nearly every case in the cohort. This systemic refinement pushed the mean Dice score to 0.896 and successfully compressed the mean spatial error (HD95) down to 6.71 mm. This bulk morphological correction required the majority of the system's interactive budget, utilizing an average workload of 18.92\%.

The quantitative mechanism driving this localized efficiency is the hardcoded biological adjacency logic embedded within the Hierarchical Topological Filtering module. In high-grade malignant gliomas, a contrast-enhancing tumor core biologically mandates surrounding peritumoral edema. By structurally enforcing this biological constraint within the CPU triage layer, the system automatically purges isolated, high-frequency false-positive core annotations occurring outside the brain parenchyma directly to background space at zero cost to the human interaction budget. This automated space clearing handles the primary burden of global false-positive reduction, which drives the Whole Tumor Dice improvement from 0.884 to 0.896 prior to human querying, leaving the human oracle completely un-diluted to focus on true boundary tracking.

\subsubsection*{Synergy of the Full Framework}
The complete UG-HITL framework demonstrates the intended algorithmic synergy. Operating strictly within the 25\% maximum dynamic budget cap, the full system successfully combines the targeted, outlier-specific rescues of the Hotspot Hunter with the zero-cost false-positive removal of topological filtering and the pervasive morphological smoothing of the gradient-snapped boundary refinement module. While the pipeline without gradient snapping achieved a marginally lower absolute HD95 (6.71 mm vs 6.85 mm), the full framework enforces strict anatomical gradient barriers. This deliberate trade-off accepts a microscopic regression in raw mathematical error to strictly prevent boundary growth into distinct healthy anatomical structures (e.g., ventricles). Thus, the finalized configuration achieves the optimal Pareto-efficient balance between realistic surgical margin correction and constrained clinical workload (18.89\%).

\section*{Discussion}
The proposed Uncertainty-Guided Human-in-the-Loop (UG-HITL) framework successfully addresses a fundamental bottleneck in the clinical deployment of deep learning for neuro-oncology: the discrepancy between high mean volumetric overlap metrics and worst-case localized topological failure \cite{cabitza2017unintended}. By transitioning the clinician from an exhaustive visual auditor to an active, localized verifier, the system establishes a Pareto-efficient pathway for safely refining surgical margins \cite{budd2021survey}. Our comprehensive evaluation on the out-of-distribution UTSW clinical cohort proves that decoupling spatial uncertainty via a Test-Time Augmentation (TTA) configuration and enforcing strict biological connectivity rules can successfully rescue severe baseline hallucinations while mathematically compressing the median interactive human workload to just 11.3\% of the target volume.

{The architectural evolution of this framework highlights the critical distinction between theoretical uncertainty performance and real-world clinical robustness. Initial in-distribution evaluations on curated in-distribution data suggested that an Evidential Deep Learning (EDL) backbone provided an optimal balance of spatial accuracy and interaction efficiency, yielding a peak Dice score of 0.990 (Table \ref{tab:idl_benchmark}). However, subsequent exposure to out-of-distribution clinical noise exposed a severe structural vulnerability in EDL's Dirichlet formulation-a phenomenon we term the "Uniform Uncertainty Trap." Because the network outputs strictly zero evidence in empty background air, the uniform probability distribution defaults to a high cumulative foreground probability ($p=0.75$) when multiple tumor sub-classes are modeled. Bypassing the standard argmax to rescue false negatives consequently causes an explosive false-positive breakdown across the entire empty background. The necessity of pivoting to a TTA underscores an essential engineering philosophy for clinical AI: one must not degrade the baseline stability of an entire clinical deployment to achieve single-pass efficiency.}

Despite its success in reducing mean spatial boundary errors, a rigorous assessment reveals specific structural and geometric limitations inherent to the framework's baseline dependencies. First, the pipeline functions strictly as an uncertainty-guided triage and refinement engine; it mathematically filters and dilates existing probability distributions but cannot generate entirely new segmentations if the primary feature extractor remains completely blind to an entire anatomical sub-region. This architectural dependency is starkly illustrated by the absolute Tumor Core Dice scores on the UTSW cohort, which remain strictly bounded by the out-of-distribution feature extraction limits of the underlying nnU-Net baseline (Table \ref{tab:final_results}). A critical direction for future work must therefore involve coupling our interactive triage framework with self-supervised domain adaptation techniques or foundational medical visual encoders to ensure robust primary feature representations prior to uncertainty extraction.

Second, the Structural tripwire and anomaly filtering mechanisms operate on localized geometric properties, leaving a dangerous vulnerability to isolated, high-confidence contralateral hallucinations. If the baseline network confidently generates a singular, contiguous mass in an incorrect, healthy brain hemisphere, the artifact satisfies the minimum volume and connected-component fragmentation thresholds, bypassing the Track A safety net. This precise vulnerability was observed in our full cohort evaluation, where a maximum post-refinement WT HD95 error of 109.15 mm was recorded (Table \ref{tab:final_results}). To mitigate this blind spot without introducing computationally heavy secondary neural classifiers, future iterations of the framework will integrate lightweight, image-level spatial verification steps directly into the CPU triage module. Implementing simple lateral hemisphere symmetry checks or evaluating the center-of-mass coordinates against standard anatomical brain atlases would provide immediate, computationally efficient failsafes capable of catching gross hemispheric violations within milliseconds, securely routing them to 100\% manual review while preserving the system's 14 to 18-second inference window.

Finally, the interaction metrics achieved must be contextualized within the constraints of the simulated Human Oracle protocol. The 11.3\% median workload represents an algorithmic upper bound of routing efficiency under the assumption of perfect clinician compliance and absolute agreement with ground truth contours. In actual clinical practice, human-computer interaction is subject to substantial inter-rater variability and cognitive friction. Clinicians are vulnerable to automation bias, potentially accepting a flawed algorithmic boundary proposal because it reduces immediate effort, or conversely, over-correcting accurate boundaries due to a lack of system trust, which would artificially inflate true interaction metrics. To bridge the gap between computational simulation and clinical reality, the next logical progression of this research involves embedding the proposed triage algorithms directly into standard hospital Picture Archiving and Communication Systems (PACS) interfaces to execute timed, institutional user studies that formally measure true cognitive reduction, interaction fatigue, and time-to-edit latency among practicing neurosurgeons.

\section*{Conclusions}

The translation of deep learning architectures into surgical-grade neuro-oncology is fundamentally bottlenecked by the silent, localized failures of automated networks and the unsustainable cognitive burden of passive human auditing. This work successfully presents an active, Uncertainty-Guided Human-in-the-Loop framework that effectively shifts the clinical paradigm from exhaustive error-hunting to targeted, efficient verification. By translating voxel-wise predictive entropy into tightly constrained, gradient-snapped geometric proposals, the system successfully bridges the gap between automated baseline constraints and surgical-grade precision. Comprehensive evaluation across 362 out-of-distribution clinical cases demonstrates that the framework aggressively targets and rescues catastrophic boundary failures—achieving a $>3.1$ mm absolute reduction in mean Tumor Core spatial error—while securely capping the median interactive clinician workload to just 11.3\% of the target volume. Crucially, the successful deployment of the structural tripwire satisfies our out-of-distribution safety requirements, {demonstrating that the system intercepts the dominant structural failure modes---fragmentation and volume collapse---under real-world clinical noise without requiring costly localized model retraining, although a characterized residual blind spot to high-confidence contralateral hallucinations remains.} Ultimately, this framework establishes a highly robust, Pareto-efficient pathway for the safe and practical deployment of clinical AI in high-stakes medical workflows.

\section*{Data availability}
The datasets used in this study are publicly available from the corresponding data repositories \cite{baid2021rsna, menze2014multimodal, reddy2026utsw}. The direct links to the datasets are as follows: UTSW Glioma Dataset: \url{https://www.cancerimagingarchive.net/collection/utsw-glioma/}, and BraTS 2023 Dataset: \url{https://www.synapse.org/Synapse:syn51514105}.

\bibliography{refs}



\section*{Author contributions statement}
\textcolor{black}{S.H. and A.K.E. conceived the study. S.H. conducted the study and performed the data curation. S.H. and A.K.E. developed the methodology and performed validation. S.H., A.Y. and A.K.E. contributed to the investigation and conceptualisation. A.K.E. provided supervision and resources. All authors reviewed and edited the manuscript.}

\section*{Additional information}
\textbf{Accession codes:} The codes will be made publicly available upon the acceptance of the manuscript; \textbf{Competing interests:} The authors declare no competing interests; \textbf{Funding:} The authors received No Funding for this work.

\end{document}